# Signal-SGN: A Spiking Graph Convolutional Network for Skeleton Action Recognition via Learning Temporal-Frequency Dynamics

Naichuan Zheng
Beijing University of Posts and Telecommunications
School of Information and Communication Engineering
Beijing, China
2022110134zhengnaichuan@bupt.edu.cn

Yuchen Du
Beijing University of Posts and Telecommunications
School of Information and Communication Engineering
Beijing, China
duyuchen@bupt.edu.cn

Hailun Xia*
Beijing University of Posts and Telecommunications
School of Information and Communication Engineering
Beijing, China
xiahailun@bupt.edu.cn

Zeyu Liang
Beijing University of Posts and Telecommunications
School of Information and Communication Engineering
Beijing, China
lzy_sfading@bupt.edu.cn

## Abstract

For skeleton-based action recognition, Graph Convolutional Networks (GCNs) are effective models. Still, their reliance on floatingpoint computations leads to high energy consumption, limiting their applicability in battery-powered devices. While energy-efficient, Spiking Neural Networks (SNNs) struggle to model skeleton dynamics, leading to suboptimal solutions. We propose Signal-SGN (Spiking Graph Convolutional Network), which utilizes the temporal dimension of skeleton sequences as the spike time steps and represents features as multi-dimensional discrete stochastic signals for temporal-frequency domain feature extraction. It combines the 1D Spiking Graph Convolution (1D-SGC) module and the Frequency Spiking Convolution (FSC) module to extract features from the skeleton represented as spiking form. Additionally, the Multi-Scale Wavelet Transform Feature Fusion (MWTF) module is proposed to extract dynamic spiking features and capture frequency-specific characteristics, enhancing classification performance. Experiments across three large-scale datasets reveal Signal-SGN exceeding state-of-the-art SNN-based methods in accuracy and computational efficiency while attaining comparable performance with GCN methods and significantly reducing theoretical energy consumption.

*Corresponding author

## 1 Introduction

Multimodal skeleton-based action recognition aims to identify human actions by analyzing motion patterns from various joint-related modalities such as joints, bones, and their temporal displacements [33]. To tackle this task, numerous methods have been proposed, including CNNs [2, 3, 11, 23, 42], LSTMs [14, 20, 21], Transformers [1, 8, 24, 32], and Graph Convolutional Networks (GCNs) [7, 10, 35, 41, 44]. Among them, ST-GCN is a pioneering method that laid the foundation for GCN-based approaches [44]. Subsequent methods, such as 2s-AGCN [35] and Dynamic GCN [48], have been proposed to achieve higher accuracy in action recognition. Lightweight models like Shift-GCN [10] and CTR-GCN [7] have optimized accuracy while significantly reducing computational costs. However, these methods are all based on artificial neural networks (ANNs), which rely on floating-point operations (FLOPs), leading to significant energy consumption that limits their practicality in energy-constrained applications.

Spiking Neural Networks (SNNs), as third-generation neural networks, use binary spike-based encoding to mimic neural dynamics. Unlike ANNs, they process information in an event-driven manner, enabling much lower energy consumption with efficient computation [15]. Leveraging this efficiency, SNNs have been applied to tasks such as image classification and point cloud estimation [38, 43, 46, 47, 53]. Spiking Graph Networks (SGNs) integrate graph convolution with SNNs, achieving up to 99% energy savings over vanilla GCNs [54]. For example, SGNs have reduced energy usage for skeleton action recognition to under 5% of GCNs while maintaining similar accuracy [52]. However, [52] treats dynamic skeletons as static images extended over spike time steps, ignoring the temporal dynamics that SNNs are designed to exploit. This increases computational cost and hinders performance, complicating training on conventional hardware. Additionally, SGNs and other SNN-based models [46, 47, 53] apply simple Global Average Pooling (GAP) across time steps, which disregards temporal dependencies in spike data, limiting the model's ability to capture temporal structure.

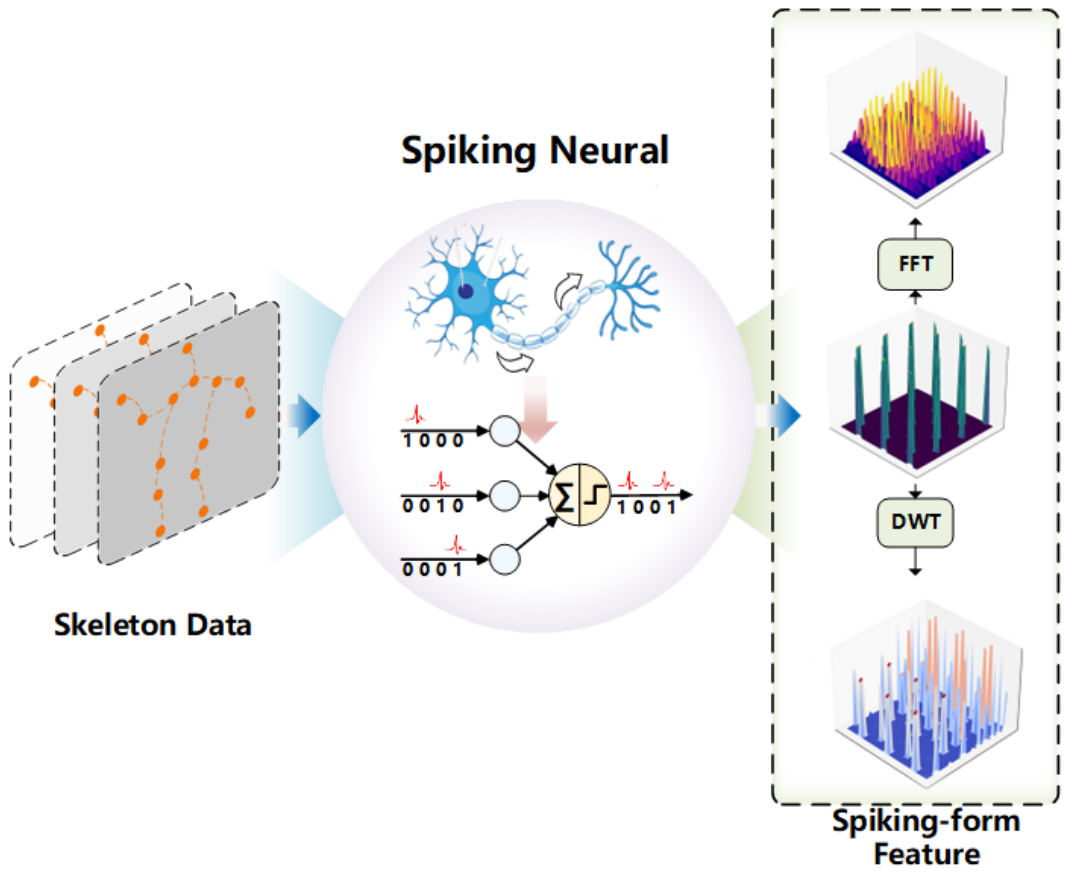

**Figure 1: Motivating perspective of Signal-SGN. Skeleton data, once encoded by the spiking neural network, becomes a sparse binary representation interpretable as a multi-dimensional discrete signal. Signal-SGN leverages this perspective to apply time-frequency analysis, revealing structured spectral-temporal patterns critical for action recognition.**

To overcome these challenges, we propose Signal-SGN for skeleton action recognition, utilizing the temporal dimension of skeleton sequences as the spiking time step and representing skeleton data as spiking-form features with multi-dimensional discrete stochastic properties, as shown in Figure1. The architecture comprises two primary modules: a one-dimensional Spiking Graph Convolution (1D-SGC) module and a Frequency Spiking Convolution (FSC) module. The 1D-SGC extracts feature from single-frame skeleton graphs within temporal sequences and represents them as the spiking-form. Meanwhile, the FSC utilizes the Fast Fourier Transform (FFT) to transform the spiking-form features of each frame into frequency-domain and leverages complex spiking convolution for advanced feature extraction. Through successive 1D-SGC-FSC layers, the features remain discrete and aperiodic. We formulate a Multi-Scale Wavelet Transform Feature Fusion (MWTF) module to address the limitations of simple GAP, which fails to effectively capture the temporal dynamics and frequency-specific characteristics of spiking-form features. The module leverages discrete wavelet transform (DWT) to decompose features into distinct frequency components and employs a spiking cross-attention mechanism to effectively fuse these components for robust feature representation. The proposed module significantly improves classification performance. Our main contributions can be summarized as follows:

- We propose a novel Signal-SGN network comprising multi-layer 1D-SGC-FSC that captures temporal-frequency domain spatial features of single-frame skeleton graphs and leverages the time-sensitive computational properties of SNNs to effectively model sequence-level temporal dynamics.
- We propose the MWTF module, which performs multi-scale temporal-frequency decomposition and integrates frequency-specific interactions to extract the temporal-frequency characteristics of spike-form features and enhance classification performance.
- Extensive experiments across three large-scale demonstrate that Signal-SGN surpasses State-of-the-Art (SOTA) SNN methods in terms of computational costs, and accuracy while achieving results comparable to GCN-based methods with significantly lower energy consumption.

## 2 Related Work

### 2.1 Skeleton-Based Action Recognition.

Deep learning methods have played a significant role in skeleton-based action recognition. Early approaches in this field achieved significant success using CNNs[2, 3, 11, 23, 42] and RNNs[14, 14, 20, 21]. Subsequent research has primarily focused on using Transformers[1, 8, 24, 32] and GCNs[7, 9, 10, 35, 44]. These numerous GCN-based methods typically involve nearly multi-layer stacks of Graph Convolutional Networks and Temporal Convolutional Networks. However, GCN-based networks are intrinsically tied to ANN architectures, which rely on floating-point computations. Such reliance leads to substantial energy consumption and fundamentally misaligns with practical requirements for energy-efficient solutions.

### 2.2 Spiking Neural Networks.

SNNs, inspired by biological neural networks, offer an energy-efficient alternative to traditional neural networks [15]. They can be implemented through two primary approaches: ANN-to-SNN conversion [16] and direct training using algorithms like STDP [6] or surrogate gradients [28]. The fundamental units, such as Leaky Integrate-and-Fire (LIF) neurons, are notable for their simplicity and biological plausibility [39]. The differential equation describes the dynamics of the LIF neuron:

$$\tau_m \frac{dV(t)}{dt} = -V(t) + R \cdot I(t), \tag{1}$$

where $V(t)$ is the membrane potential, $\tau_m$ is the time constant indicating potential decay, $R$ is the membrane resistance, and $I(t)$ is the input current. When $V(t)$ reaches the threshold $V_{\text{th}}$, the neuron fires a spike and resets to $V_{\text{reset}}$. The LIF neuron model accurately replicates the spiking dynamics characteristic of neurons within SNNs. Inspired by ANN architectures, corresponding SNN models have been proposed and applied to tasks such as image classification, event-driven processing, and EEG analysis[4, 5, 31]. Among these, various spiking-based Transformers have played significant roles[46, 47, 53]. Meanwhile, [52] introduces SGN for skeleton-based action recognition for the first time, achieving promising results. However, [52] treats the skeleton sequence data as static and artificially extends them along the spike time step dimension, which increases storage and computational demands. Additionally, SGNs fail to capture the discrete and non-periodic nature of spike-form features and rely on simple GAP to reduce the spike time step dimension before the classification head, limiting classification performance and accuracy.

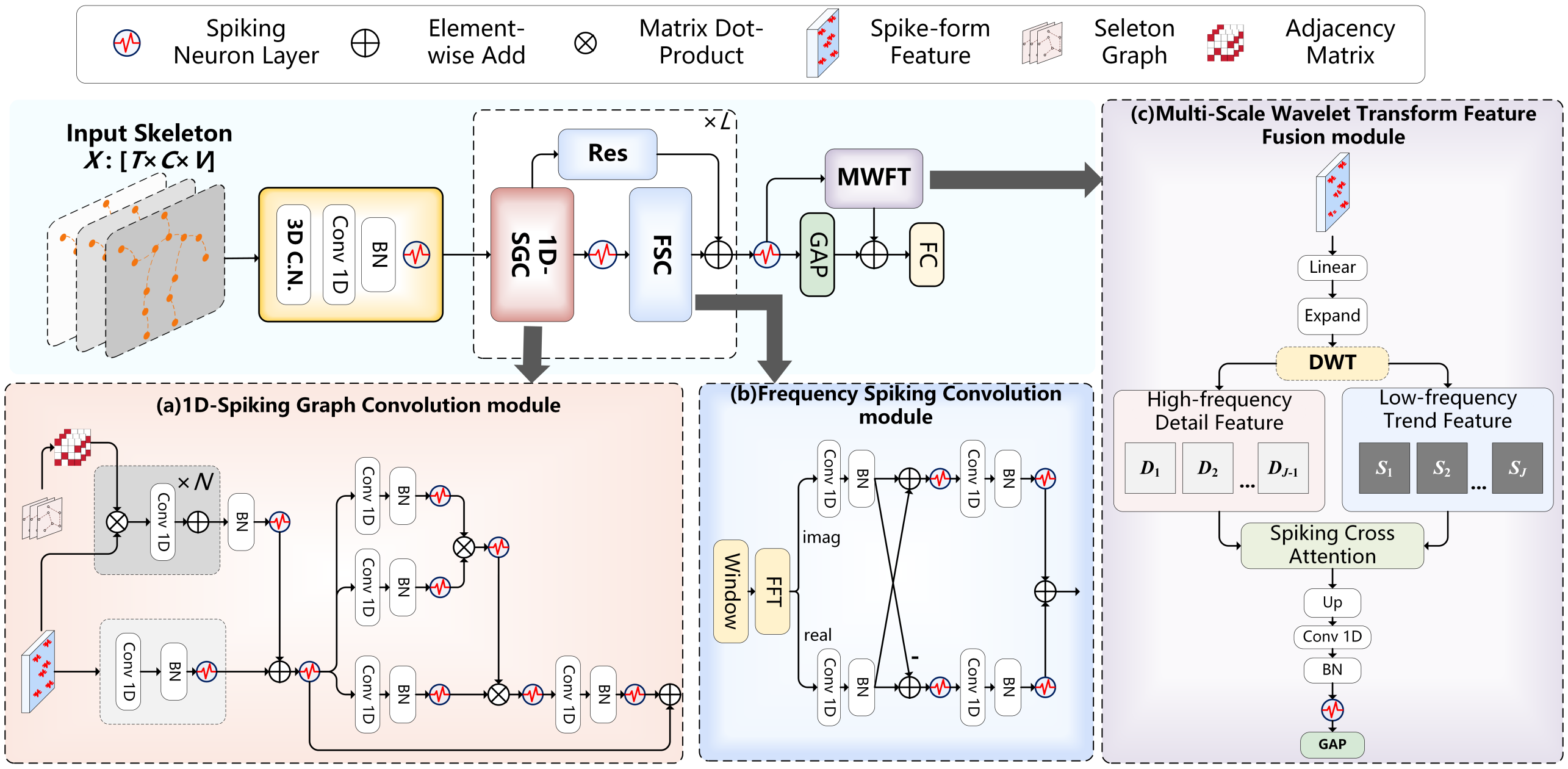

**Figure 2: Overview of the proposed Signal-SGN framework for skeleton-based action recognition. The model processes skeleton sequences through four main stages: a 3D Coordinate Normalization (3D C.N.) to stabilize input representations, a 1D-Spiking Graph Convolution (1D-SGC) module to capture topological dependencies in the spiking domain, a Frequency Spiking Convolution (FSC) module that extracts frequency-aware features via FFT and complex convolution, and a Multi-Scale Wavelet Transform Feature Fusion (MWTF) module that performs multi-level decomposition and integration using discrete wavelet transform and spiking cross-attention. Final classification is conducted through global average pooling (GAP) and a fully connected (FC) layer.**

### 2.3 Frequency Domain Analysis.

Frequency-domain analysis has become an increasingly important tool across various deep learning architectures for enhancing robustness and representational capacity[49]. The Fourier Transform (FT) and Wavelet Transform (WT) have been extensively adopted in CNNs for image compression and reconstruction [22], in RNNs for modeling sequential dynamics such as speech [18], and in GCNs for processing graph signals [50]. Recent work also demonstrates that adaptive frequency-based token mixing can replace costly attention mechanisms while maintaining global context modeling capabilities [37]. Moreover, in camouflaged object detection, frequency-spatial entanglement learning has proven effective in mitigating the limitations of local spatial features [36]. In the context of Vision Transformers, frequency-guided representations have also shown to improve generalization and robustness by modulating attention with spectral priors [30].

In the context of SNNs, skeleton sequences are encoded as discrete multi-dimensional spike signals—akin to stochastic binary signals. Exploring their frequency characteristics offers promising potential for interpretable analysis and enhanced recognition. However, frequency-domain analysis for SNN-based skeleton action recognition remains underexplored and merits further attention.

## 3 Method

This section details the architecture of the proposed Signal-SGN, as illustrated in Figure 2. Given an input skeleton sequence, the model first applies a 3D Coordinate Normalization (3D C.N.) to standardize joint coordinates and enhance spiking encoding consistency. Then, the 1D-SGC module captures spatial dependencies of joints through spiking-based graph convolution. To further exploit frequency-domain information, FSC module transforms features via FFT and applies complex convolution for frequency-aware representation learning. Subsequently, the MWTF module decomposes spike-form features into multi-resolution frequency components using discrete wavelet transform, followed by cross-attentive fusion. The final prediction is made through GAP and a fully connected(FC) layer.

### 3.1 3D C.N.

As shown in Figure 3, 3D skeleton data spans multiple spatial quadrants and includes both positive and negative coordinate values. This characteristic introduces challenges for SNNs, which are inherently designed for non-negative numerical inputs. In particular, the presence of negative values can disrupt rate-based spike encoding and lead to severe information loss due to binarization effects.

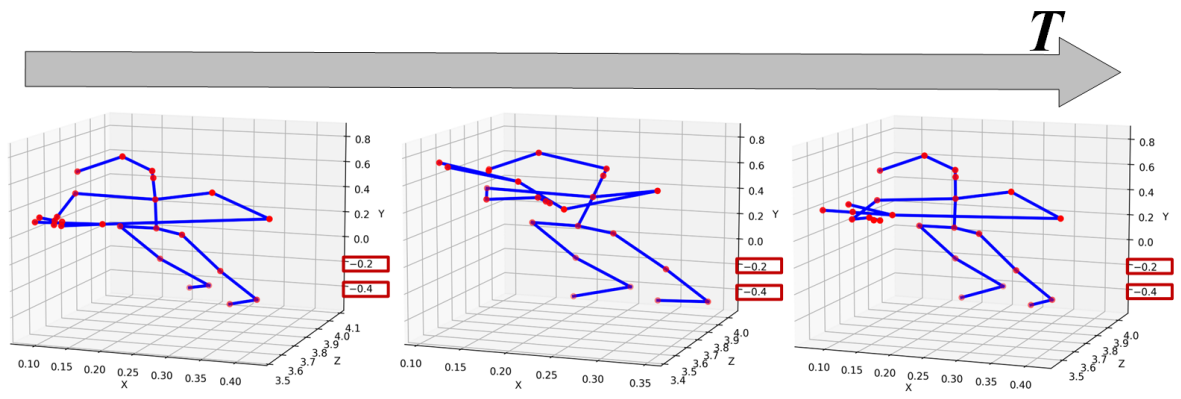

**Figure 3: 3D skeleton visualization showing both positive and negative coordinate values.**

To address this, we apply a per-channel normalization that maps each spatial coordinate to the range $[0, 1]$. Given a skeleton sequence $X \in \mathbb{R}^{C\times T\times V}$, where $C = 3$ denotes the spatial dimensions $(x, y, z)$, $T$ is the number of frames, and $V$ is the number of joints, we normalize each coordinate channel $c \in \{x, y, z\}$ independently across the full sequence utilizing:

$$c' = \frac{c - \min(c)}{\max(c) - \min(c)} \tag{2}$$

where $\min(c)$ and $\max(c)$ are computed over all joints and time steps for each coordinate channel. This process preserves the relative geometric structure while ensuring compatibility with spike-based encoding.

The normalized skeleton input is then processed by a one-dimensional convolution (Conv1D), followed by batch normalization (BN) and a spiking neuron (SN) layer. This sequence forms the initial encoding stage of Signal-SGN, and the resulting feature is denoted as $X_{\mathrm{i}}$.

## 3.2 1D-SGC Module

In the 1D-SGC module, inspired by GCN, we integrate topological information by normalizing adjacency-based weights to aggregate attributes from neighboring nodes, as shown in Figure 2(a). The skeleton can be represented as a graph $\mathcal{G}(V, E)$, where the joints form a set of vertices $V$ and the bones represent the edges $E$. This structure can be expressed as an adjacency matrix $A \in \mathbb{R}^{V\times V}$, where $A_{i,j} = 1$ if joints $i$ and $j$ are physically connected, otherwise $A_{i,j} = 0$. To incorporate topological relationships into computations, the normalized adjacency matrix $S$ is computed as follows:

$$S = \Delta^{-\frac{1}{2}}\widetilde{A}\Delta^{-\frac{1}{2}}, \tag{3}$$

where $\widetilde{A} = A+I$ is the adjacency matrix with added self-connections $I$, and $\Delta$ is the degree matrix of $\widetilde{A}$. This normalization ensures that the influence of each node is appropriately scaled based on its connectivity, facilitating the effective integration of topological information into the model. Subsequently, we apply Conv1d and BN along channel dimension to extract features for each skeleton node within each spike time step, effectively aggregating features across the skeleton joints dimension.

$$G_{\mathrm{i}} = \mathrm{BN}\left[\sum_{n=0}^{N} \mathrm{Conv1d}\left(X_{\mathrm{i}} * S^{n}\right)\right], \tag{4}$$

in which $S^n$ represents a set of matrices decomposed from $S$, $N$ refers to the number of different relational matrices in the decomposition, and $*$ denotes matrix multiplication. The resulting output $G_{\mathrm{i}} \in \mathbb{R}^{T\times D\times V}$, where $D$ represents the hidden channel dimension.

To leverage the temporal dynamic characteristics of SNNs, we map the inherent temporal dimension of the features to the spike time steps of the SNNs and input the features into spiking neurons ($SN$). The Algorithm 1 outlines the detailed process. provided in the Appendix C.1. For compactness, we define the operation $SN[\mathrm{BN}(\mathrm{Conv1d}(\cdot))]$ as $\Phi(\cdot)$. The updated formula becomes:

$$G_{\mathrm{o}}^{(0)} = SN\left(G_{\mathrm{i}}\right) + \Phi\left(X_{\mathrm{i}}\right). \tag{5}$$

Building upon the approach proposed in [53], we integrate the spike self-attention (SSA) mechanism after the graph convolution operation to enhance feature representation. The formula is as follows:

$$Q^{(0)} = \Phi_Q(G_{\mathrm{o}}^{(0)}), K^{(0)} = \Phi_K(G_{\mathrm{o}}^{(0)}), V^{(0)} = \Phi_V(G_{\mathrm{o}}^{(0)}), \tag{6}$$

$$G^{(0)} = G_{\mathrm{o}}^{(0)} + \Phi\left(Q^{(0)}K^{(0)T}V^{(0)} * s\right), \tag{7}$$

where $s$ is the scaling factor to control the large value of the matrix multiplication result.

## 3.3 FSC Module

Following the 1D-SGC module, the features are re-transformed by $SN$ into spiking-form feature—discrete, aperiodic signal sequences encoded as binary values (0s and 1s). To further extract meaningful frequency-domain feature, we propose the FSC module, as shown in Figure 2(b).

First, this module leverages a learnable window function $w_V(v) \in \mathbb{R}^V$ along the joint dimension to mitigate spectral leakage and enhance frequency analysis. Unlike fixed windows, $w_V$ assigns adaptive weights to each joint, acting as a soft spatial gate before the frequency transform and applied element-wise to the input feature map $G^{(0)}$ as:

$$F^{(0)} = G^{(0)} \cdot w_V(v), \tag{8}$$

where $F^{(0)}$ denotes the windowed features passed to the frequency encoder, and $\cdot$ denotes broadcasting over the joint dimension.

We perform a Fast Fourier Transform (FFT) along the joint dimension $V$ to project the input feature $F_{\mathrm{i}}^{(0)}$ into the frequency domain. The resulting complex-valued representation is separated into real and imaginary components, denoted as $F_{\mathrm{R}}^{(0)}$ and $F_{\mathrm{I}}^{(0)}$, which are subsequently processed by parallel branches. This transformation enables the model to extract interpretable spatial-frequency patterns across joints, facilitating enhanced action representation. To process the complex-valued FFT outputs, we propose a spiking complex convolution module that preserves and utilizes amplitude and phase information, enhancing the network's ability to capture detailed frequency-domain features:

$$F_{\mathrm{o,R}}^{(0)} = \mathrm{BN}[\mathrm{Conv1d}(F_{\mathrm{R}}^{(0)})], F_{\mathrm{o,I}}^{(0)} = \mathrm{BN}[\mathrm{Conv1d}(F_{\mathrm{I}}^{(0)})], \tag{9}$$

$$F_{\mathrm{o},u}^{(0)} = \mathrm{SN}\left(F_{\mathrm{o,R}}^{(0)} \pm_u F_{\mathrm{o,I}}^{(0)}\right), \quad u \in \{1, 2\}. \tag{10}$$

where $\pm_k$ represents addition $(+)$ when $u = 1$, and subtraction $(-)$ when $u = 2$. The fused output is then computed as:

$$F^{(0)} = \Phi\left(F_{\mathrm{o},1}^{(0)}\right) + \Phi\left(F_{\mathrm{o},2}^{(0)}\right). \tag{11}$$

The output $F^{(0)}$ represents the extracted frequency-domain features, serving as an essential input for the subsequent layers of the network. Then a complete 1D-SGC-FSC process can then be expressed as:

$$X^{(l)} = F^{(l)} + G^{(l)} + \text{res}\left(G^{(l-1)}\right) \tag{12}$$

where $X^{(l)}$ denotes the output of the $l$-th layer, combining the frequency-domain features $F^{(l)}$, temporal-domain features $G^{(l)}$, and a residual connection from the previous layer $\text{res}\left(G^{(l-1)}\right)$.

## 3.4 Multi-Scale Wavelet Transform Feature Fusion module

We formulate the MWTF module to capture the temporal dynamics and frequency-specific characteristics of spiking-form features, with its outputs fused with backbone outputs to enhance classification performance, as illustrated in Figure 2(c).

We first construct filters $\Lambda_0$, $\Lambda_1$, $\Gamma_0$, and $\Gamma_1$ based on Legendre polynomials $P_k(x)$ (details are provided in Appendix B) to decompose spiking-form features. The parameter $k$ specifies the number of frequency coefficients extracted, influencing the spectral resolution. We apply a linear transformation to the spiking features from the final layer $L$ of the backbone network to align the output tensor with the $k$ coefficients, ensuring compatibility with the filters.

$$X_{\text{down}} = \text{LN}\left(SN\left(X^{(L)}\right)\right) \in \mathbb{R}^{T\times k\times V}. \tag{13}$$

To ensure compatibility with the wavelet transform, we first check whether the temporal dimension $T$ is a power of two. If it is not, we extend the input spike-form feature $X_{\text{down}}$ by concatenating extra frames, denoted as $X_{\text{extra}}$, which are duplicated copies of the beginning frames. The total number of time steps is then increased to the nearest power of two. Formally, the number of extra frames added is calculated as $2^{\lceil \log_2 T\rceil} - T$, and the final extended input becomes $\widetilde{X} = [X_{\text{down}}, X_{\text{extra}}]$ along the temporal axis.

The extended input $\widetilde{X}$ is then recursively decomposed by a set of wavelet filter matrices, namely $\Lambda_0, \Lambda_1, \Gamma_0$, and $\Gamma_1$, yielding two types of outputs at each level: detail coefficients $D_j$, which capture high-frequency, localized temporal features, and scaling coefficients $S_j$, which preserve the lower-frequency trend components, as shown in Figure 4. A complete iterative decomposition process is provided in Algorithm 2 in Appendix C.2.

To enhance the interaction between detail coefficients $D_j$ and scaling coefficients $S_j$ across decomposition levels $j$, we first upsample them to the original temporal resolution $T$. Then, a spiking cross-attention mechanism is applied to capture the mutual dependencies between high-frequency and low-frequency components:

$$\hat{D}_j = \left(\Phi_Q(D_j)\cdot\Phi_K(D_j)^T\right)\Phi_V(S_j)\cdot s, \tag{14}$$

$$\hat{S}_j = \left(\Phi_Q(S_j)\cdot\Phi_K(S_j)^T\right)\Phi_V(D_j)\cdot s, \tag{15}$$

where $s$ is a learnable scaling factor. The scaling coefficient from the final level $S_J$ is only upsampled (without interaction) and denoted as $\hat{S}_J$. All outputs $\{\hat{D}_j, \hat{S}_j\}_{j=0}^{J-1} \cup \{\hat{S}_J\}$ are aggregated and passed through spiking activation and global average pooling to yield the multi-level output:

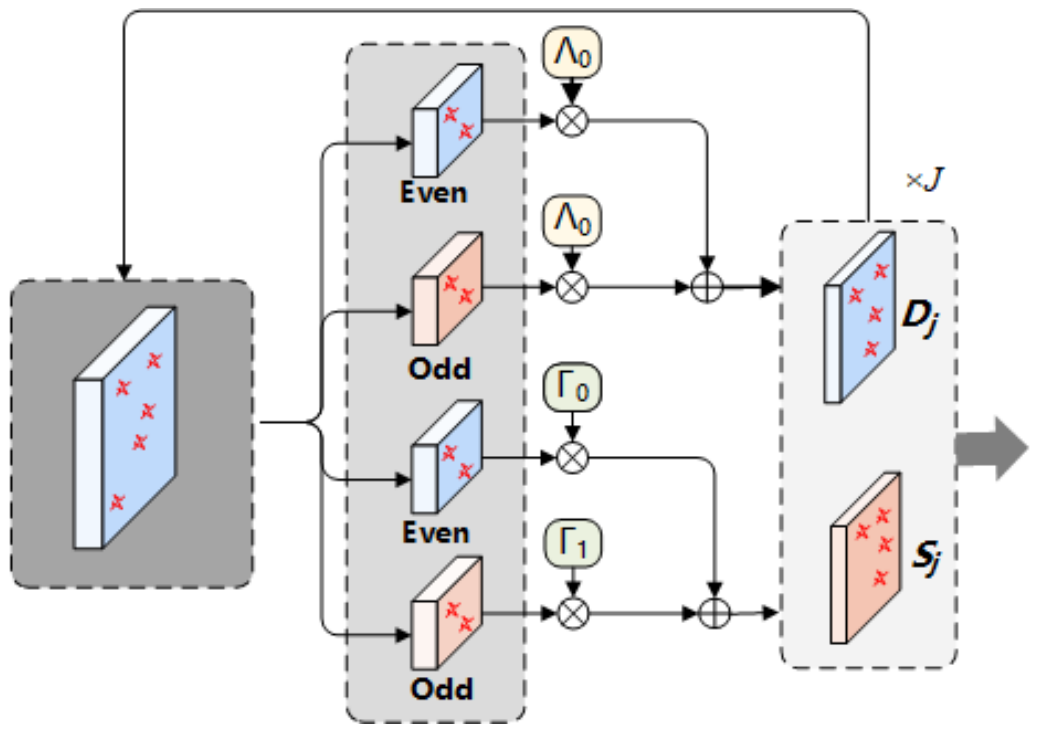

**Figure 4: The process of DWT**

$$\hat{X} = \text{GAP}\left(\Phi\left(\sum_{j=0}^{J-1}\hat{D}_j + \sum_{j=0}^{J}\hat{S}_j\right)\right). \tag{16}$$

The resulting feature $\hat{X}$ is then fused with the SN-processed output of the final backbone layer $X^{(L)}$, and the fused representation is sent to a fully connected layer for classification:

$$\hat{y} = \text{FC}\left(\text{GAP}\left(SN\left(X^{(L)}\right)\right) + \beta\hat{X}\right), \tag{17}$$

where $\beta$ is a learnable parameter balancing the contribution of temporal and spectral features. Cross-entropy loss is used for optimization during training.

# 4 Experiments

To demonstrate the effectiveness of Signal-SGN, we conduct skeleton-based action recognition experiments on three widely used datasets: NTU RGB+D[34], NTU RGB+D 120[25], and NW-UCLA[40]—under standard evaluation protocols including Cross-Subject (Xs), Cross-View (Xv), and Cross-Setup (Xt), covering diverse actions, subjects, and viewpoints. We compare our model against strong ANN and SNN baselines and perform ablation studies to evaluate the contribution of each module. The following sections detail the experimental setup, including dataset settings, hyperparameter configurations, comparisons with state-of-the-art methods, and ablation results.

## 4.1 Model Implementation and Configuration

Our model is implemented using PyTorch[1] and SpikingJelly[2], and all experiments are conducted on four NVIDIA V100 GPUs. The backbone network consists of 4 layers with input and output channel sizes of 3 and 64, 64 and 64, 64 and 128, 128 and 256. We use SGD optimizer with a learning rate of 0.1, decay rate of 0.1 at step 110, batch size of 64, training for 150 epochs with a dropout rate of 0.3, weight decay of 0.0001, and auxiliary loss weight $\omega$ set to 0.03.

## 4.2 Comparison with the State-of-the-Art

Several SOTA methods adopt multi-stream fusion strategies for improved performance [7, 10, 35]. To ensure a fair comparison, we

[1]https://github.com/rwightman/pytorch-image-models
[2]https://github.com/fangwei123456/spikingjelly

**Table 1: Comparison of Signal-SGN with prior ANN- and SNN-based models on NTU RGB+D, NTU RGB+D 120, and NW-UCLA datasets. We report classification accuracy (%), model size (Param.), computational cost (FLOPs and SOPs), and energy consumption (Power).**

| Model (ANN/SNN) | Param. (M) | FLOPs (G) | SOPs (G) | Power (mJ) | NTU RGB+D Xs(%) | NTU RGB+D Xv(%) | NTU RGB+D 120 Xs(%) | NTU RGB+D 120 Xt(%) | NW-UCLA (%) |
|---|---|---|---|---|---|---|---|---|---|
| Part-aware LSTM [12] | - | - | - | - | 62.9 | 70.3 | 25.5 | 26.3 | - |
| Spatio-Temporal LSTM [26] | - | - | - | - | 61.7 | 75.5 | 55.7 | 57.9 | - |
| ST-GCN [44] | 3.10 | 3.48 | - | 16.01 | 81.5 | 88.3 | 70.7 | 73.2 | - |
| 2S-AGCN [35] | 3.48 | 37.32 | - | 182.87 | 88.5 | 95.1 | 82.5 | 84.3 | - |
| Shift-GCN [10] | - | 10.00 | - | 46.00 | 90.7 | 96.5 | 85.3 | 86.6 | 94.6 |
| MS-G3D [27] | 3.19 | 48.88 | - | 239.51 | 91.5 | 96.2 | 86.9 | 88.4 | - |
| CTR-GCN [7] | 1.46 | 7.88 | - | 36.25 | 92.4 | 96.8 | 88.9 | 90.6 | 96.5 |
| Spikformer [53] | 4.78 | 24.07 | 1.69 | 2.17 | 73.9 | 80.1 | 61.7 | 63.7 | 85.4 |
| Spike-driven Transformer [47] | 4.77 | 23.50 | 1.57 | 1.93 | 73.4 | 80.6 | 62.3 | 64.1 | 83.4 |
| Spike-driven Transformer V2 [46] | 11.47 | 38.28 | 2.59 | 2.91 | 77.4 | 83.6 | 64.3 | 65.9 | 89.4 |
| Spiking Wavelet Transformer [13] | 3.24 | 19.30 | 1.48 | 2.01 | 74.7 | 81.2 | 63.5 | 64.7 | 86.7 |
| MK-SGN [52] | 2.17 | 7.84 | 0.68 | 0.614 | 78.5 | 85.6 | 67.8 | 69.5 | 92.3 |
| Signal-SGN (Joint) | 1.74 | 1.62 | 0.314 | 0.372 | 80.5 | 87.7 | 69.2 | 72.1 | 92.7 |
| Signal-SGN (Bone+Joint) | 1.74 | 3.24 | 0.628 | 0.744 | 82.5 | 89.2 | 71.3 | 74.2 | 93.1 |
| **Signal-SGN (4 ensemble)** | **1.74** | **6.48** | **1.28** | **1.488** | **86.1** | **93.1** | **75.3** | **77.9** | **95.9** |

Param. represents the model size for the network itself. For ANN methods, FLOPs, power, and accuracy are reported under standard protocols. For SNN methods, SOPs and energy estimates are computed based on spike-driven inference.

employ the same four-stream fusion protocol in Signal-SGN, incorporating joint, bone, joint motion, and bone motion modalities. We evaluate Signal-SGN against both ANN and SNN baselines on NTU RGB+D, NTU RGB+D 120, and NW-UCLA. Classification accuracy, model size, FLOPs, synaptic operations(SOPs), and estimated energy consumption are summarized in Table 1, with calculation details provided in Appendix A. On the NTU RGB+D and NTU RGB+D 120 datasets, all sequences are downsampled to 16 frames. Compared to prior SNNs, Signal-SGN (Joint) achieves 80.5% accuracy on NTU RGB+D (Xs) while requiring only 0.372 mJ of energy—an 82.9% reduction compared to Spikformer's 2.17 mJ, and a 98.9% reduction compared to the ANN-based MS-G3D (239.51 mJ). Notably, our four-stream fused model further boosts accuracy to 86.1% (Xs) and 93.1% (Xv) on NTU RGB+D and 75.3% (Xs) / 77.9% (Xt) on NTU RGB+D 120, while still keeping energy usage at 1.488 mJ, which is over 96% lower than CTR-GCN (36.25 mJ) and 99.4% lower than MS-G3D. This demonstrates Signal-SGN's superior trade-off between performance and efficiency. On the NW-UCLA, Signal-SGN achieves a competitive 95.9% accuracy, only 0.6% below CTR-GCN, while using less than 5% of its theoretical energy—further validating the model's generalizability and low-cost design.

## 4.3 Ablation Studies

To evaluate the contributions of the primary components in Signal-SGN, we conducted ablation studies on the NTU-RGB+D (Xs) dataset based on joint data, with all sequences resized to 16 frames.

*4.3.1 Module-wise Ablation Analysis.* We first module-wise ablation analysis to evaluate the effectiveness of the proposed components in Signal-SGN. As summarized in Table 2, the base model with only the 1D-SGC module achieves 69.7% accuracy. Adding the FSC significantly boosts performance to 77.0%, indicating the advantage of incorporating frequency-domain information. Further introducing the MWTF module improves accuracy to 79.3%, demonstrating its ability to capture fine-grained temporal-frequency dynamics. A slight improvement is also observed when applying 3D coordinate normalization across different configurations, suggesting it can further stabilize input distributions and facilitate downstream spiking representation learning. Overall, these results confirm the progressive effectiveness of the proposed modules, with MWTF contributing most significantly on top of FSC.

**Table 2: Module-wise ablation study on NTU-RGB+D (Xs). We report model size, computational cost, and top-1 accuracy. Spikformer is included as a baseline for external comparison.**

| 3D C.N. | 1D-SGC | FSC | MWTF | Param. (M) | FLOPs / SOPs (G) | Acc (%) |
|---|---|---|---|---|---|---|
| Spikformer (baseline)[53] | | | | 4.78 | 24.07 / 1.69 | 73.9 |
| - | ✓ | - | - | 1.03 | 1.12 / 0.252 | 69.7 |
| - | ✓ | ✓ | - | 1.65 | 1.60 / 0.312 | $77.0^{+7.3}$ |
| - | ✓ | ✓ | ✓ | 1.74 | 1.62 / 0.314 | $79.3^{+2.3}$ |
| ✓ | ✓ | - | - | 1.03 | 1.12 / 0.252 | 70.5 |
| ✓ | ✓ | ✓ | - | 1.65 | 1.60 / 0.312 | $78.3^{+7.8}$ |
| ✓ | ✓ | ✓ | ✓ | 1.74 | 1.62 / 0.314 | **80.5**$^{+2.2}$ |

*4.3.2 Ablation study on the FSC module.* To validate the effectiveness of the proposed FSC module in enhancing spiking feature representation, we conduct comprehensive ablation studies, as shown in Table 3. Introducing only the real or imaginary branch improves performance from 70.5% to 72.3% and 72.1%, respectively, suggesting that both components contribute meaningfully. Combining them without interaction ($u$) further increases accuracy to 73.6% (+1.5%), confirming their complementarity. Enabling single interaction branches—either $u = 1$ or $u = 2$—leads to 75.4% and 74.9%, respectively, while using both ($u = \{1, 2\}$) achieves the best performance at 78.3%, yielding a +3.4% gain over the best single-branch configuration and +7.8% over the baseline. Replacing the learnable window with a fixed Hamming window slightly reduces accuracy to 74.9%, and removing the window entirely yields 77.8%, showing the window's limited but non-negligible contribution. These results demonstrate that the core benefit of FSC arises from complex-valued modeling and frequency-domain interaction mechanisms. To further understand the role of the learnable window, we conduct a visualization analysis across FSC layers, as detailed in Appendix D.2.

**Table 3: Ablation study on the FSC module.**

| Configuration | $u$ Interaction | Window | Acc (%) |
|---|---|---|---|
| Without FSC | – | – | 70.5 |
| Real branch only | - | Learnable | $72.3^{+1.8}$ |
| Imaginary branch only | - | Learnable | $72.1^{-0.2}$ |
| Real + Imag (no fusion) | - | Learnable | $73.6^{+1.5}$ |
| Only $u = 1$ branch | $u = 1$ | Learnable | $75.4^{+1.8}$ |
| Only $u = 2$ branch | $u = 2$ | Learnable | $74.9^{-0.5}$ |
| Full FSC | $u = 1, 2$ | Learnable | $\mathbf{78.3}^{+3.4}$ |
| Full FSC | $u = 1, 2$ | Hamming | $78.0^{-0.3}$ |
| Full FSC | $u = 1, 2$ | None | $77.8^{-0.2}$ |

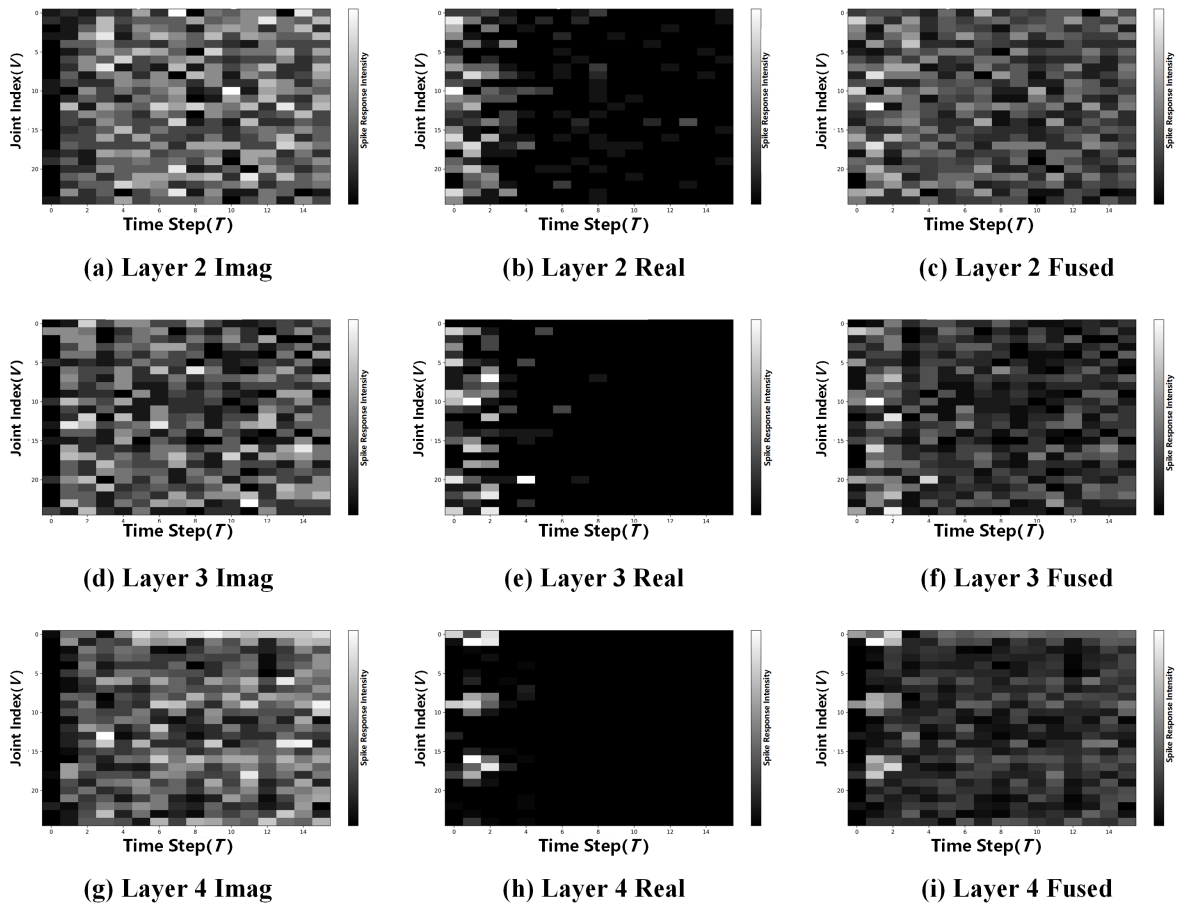

**Figure 5: Spike response visualizations of FSC for the "drinking water" action sample.**

*4.3.3 Qualitative Visualization of FSC module.* To better understand the spiking neural behavior of our model, we visualize the joint-time ($V \times T$) spiking heatmaps from a representative "drinking water" sample, as shown in Figure 5. For each of the four FSC layers, we display activations from the real, imaginary, and fused branches, where grayscale intensity encodes the summed spike responses over the channel dimension. The fused features consistently exhibit smoother and more structured spatiotemporal patterns, indicating that combining real and imaginary branches enhances feature stability and expressiveness. In contrast, the real branch shows sparser, localized firing, while the imaginary branch captures more dispersed activity. Importantly, in the final FSC layer, joints associated with the right arm—specifically the right elbow (10), wrist (11), and hand (12)—exhibit prominent activations, aligning with the expected motion of lifting a cup. Symmetric activity is also observed in joints 6–8 (left arm), reflecting bilateral arm coordination. These observations validate that FSC learns semantically meaningful joint dynamics, and that the fused representation captures discriminative patterns critical for accurate action recognition.

**Table 4: Comparison of experimental results under different MWTF settings ($k$ and $J$)**

| MWTF | $k$ | $J$ | Param. (M) | FLOPs/SOPs (G) | Acc (%) |
|---|---|---|---|---|---|
| - | - | - | 1.65 | 1.60/0.312 | 78.3 |
| ✓ | 4 | 1 | 1.68 | 1.61/0.313 | $78.9^{+0.6}$ |
| ✓ | 8 | 1 | 1.73 | 1.62/0.314 | $79.8^{+0.9}$ |
| ✓ | 8 | 2 | 1.74 | 1.62/0.315 | $80.3^{+0.5}$ |
| ✓ | 8 | 3 | 1.74 | 1.62/0.317 | $\mathbf{80.5}^{+0.2}$ |
| ✓ | 16 | 3 | 1.83 | 1.64/0.320 | $80.5^{-}$ |

**Table 5: Ablation study on spiking cross-attention and inclusion of $S_J$ in MWTF ($k$=8, $J$=3).**

| Setting | $\hat{D}_j$ Branch | $\hat{S}_j$ Branch | $\hat{S}_J$ | Acc. (%) |
|---|---|---|---|---|
| None (Baseline) | – | – | – | 78.3 |
| Only $\hat{D}_j$ | ✓ | – | – | $78.9^{+0.6}$ |
| Only $\hat{S}_j$ | – | ✓ | – | $78.7^{-0.2}$ |
| Only $S_J$ | – | – | ✓ | $78.4^{-0.3}$ |
| $\hat{D}_j + \hat{S}_j$ | ✓ | ✓ | - | $79.8^{+0.4}$ |
| $\hat{S}_j + S_J$ | – | ✓ | ✓ | $79.2^{-0.6}$ |
| $\hat{D}_j + S_J$ | ✓ | - | ✓ | $79.4^{+0.2}$ |
| Full (Ours) | ✓ | ✓ | ✓ | $\mathbf{80.5}^{+1.3}$ |

*4.3.4 Ablation study on the MWTF module.* To further analyze the effect of the MWTF module, we conduct an ablation study on its two key hyperparameters: $k$ (number of frequency coefficients) and $J$ (number of decomposition levels). The results, presented in Table 4, evaluate the trade-off between model complexity and classification accuracy. Starting from the baseline model without MWTF (78.3%), incorporating low-order configurations ($k$=4, $J$=1) leads to a modest gain (+0.6%), with the best performance (80.5%) achieved at $k$=8, $J$=3. This validates the importance of multi-level frequency decomposition. However, further increasing $k$ to 16 brings no improvement, indicating saturation.

To further dissect MWTF's contributions, we ablate the two spiking cross-attention branches ($\hat{D}_j$, $\hat{S}_j$) and the use of the final approximation coefficient $\hat{S}_J$. Results in Table 5 show that each component brings incremental benefits, with $\hat{D}_j$ Branch alone improving performance to 78.9%, and the full configuration reaching 80.5%. To complement the analysis, we also investigate the evolution of the learnable fusion weight $\beta$ training. The results, including its stabilization behavior, are provided in Appendix D.1.

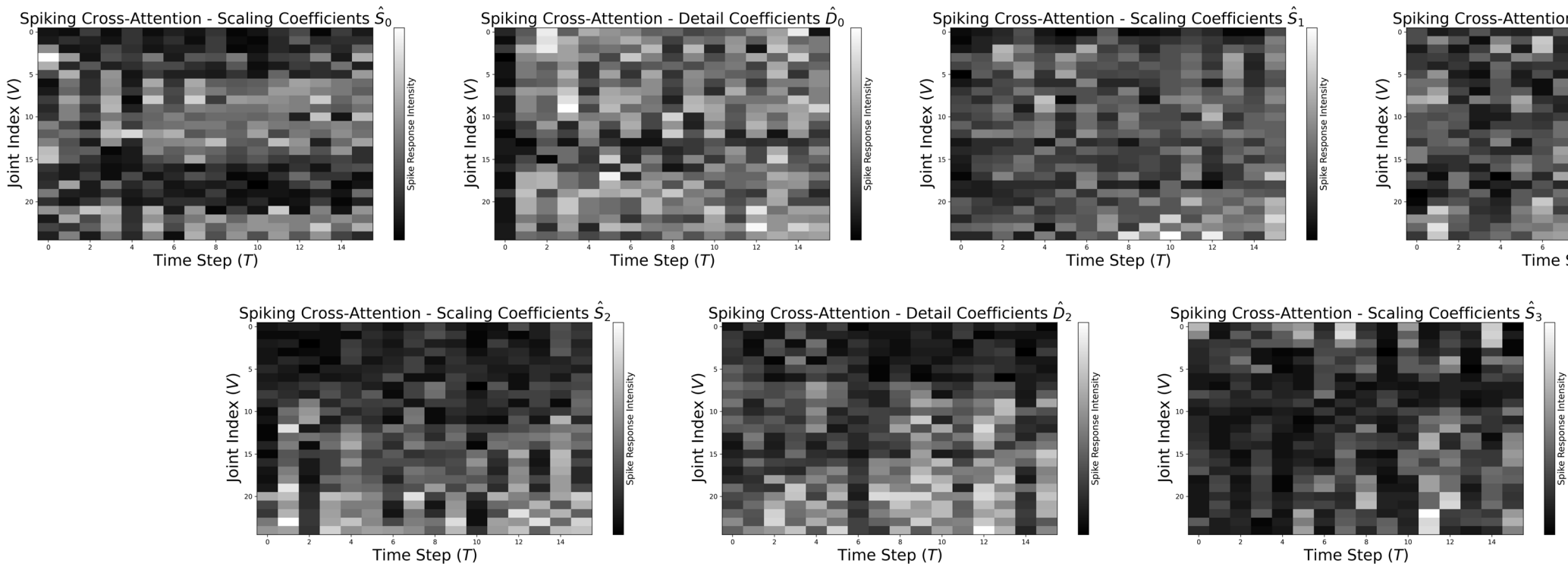

**Figure 6: Visualization of upsampled detail coefficients $\hat{D}_j$ and scaling coefficients $\hat{S}_j$ across levels $j = 0, 1, 2$, along with the final approximation coefficient $S_3$ in MWTF. Each heatmap shows the spike response intensity over joints and time steps, revealing the multi-scale temporal characteristics captured by the spiking cross-attention mechanism.**

*4.3.5 Interpretability Analysis of MWTF via CAM Visualization.* To further validate the interpretability of the MWTF module, we visualize the joint-level Class Activation Maps (CAMs) derived from its output features for three representative actions: *Drink Water*, *Stand Up*,*Reach into Pocket*,*Phone Call*,*Tear up Paper* and *Throw*(Figure 7). These CAMs highlight the spatial regions most responsive in the model's representation. In the *Drink Water* example, high activation appears near the right wrist and elbow, matching the expected movement pattern. In *Stand Up*, the model emphasizes lower body joints, while *Reach into Pocket* shows localized attention around the hip and wrist regions. These joint-specific activations are consistent with the underlying motion characteristics of each action. The results demonstrate that MWTF's multi-scale frequency decomposition effectively captures action-relevant joint responses, confirming the strength of time-frequency analysis in spiking skeleton-based action recognition.

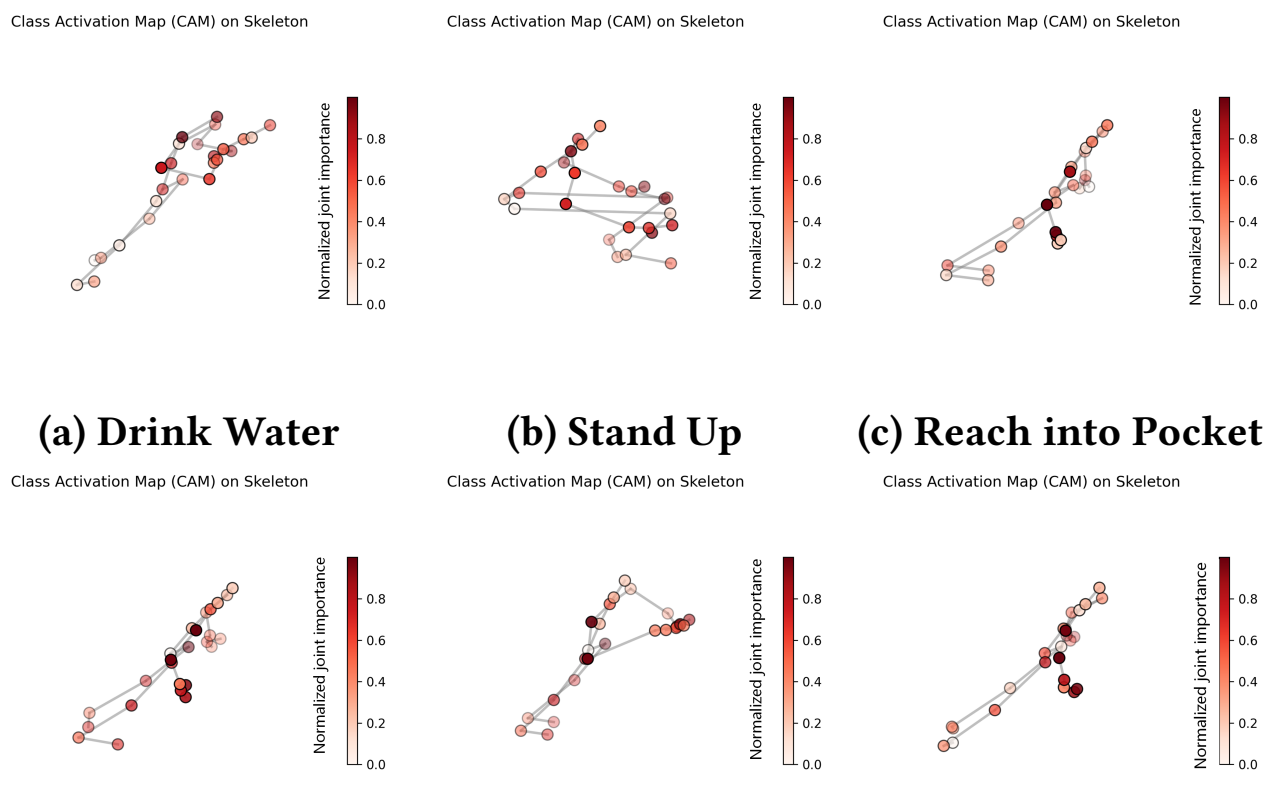

**(a) Drink Water** **(b) Stand Up** **(c) Reach into Pocket**

**(d) Phone Call** **(e) Tear Up Paper** **(f) Throw**

**Figure 7: Class Activation Maps (CAMs) based on MWTF outputs for three representative actions. Warmer colors indicate higher joint-level importance.**

*4.3.6 Qualitative Visualization of Multi-Scale Coefficients.* To assess the impact of multilevel decomposition and spiking cross-attention in MWTF, we visualize the upsampled detail $\hat{D}_j$ and scaling $\hat{S}_j$ coefficients ($j = 0, 1, 2$), along with the final approximation coefficient $S_3$, for a representative *drinking water* action (Figure 6). The heatmaps reveal that $\hat{D}_j$ emphasizes localized, high-frequency joint activations, while $\hat{S}_j$ captures smoother, global trends. Particularly, $S_3$ exhibits temporally stable responses, reflecting semantic patterns preserved through hierarchical decomposition. The complementarity between $\hat{D}_2$ and $\hat{S}_2$ further demonstrates the effectiveness of spiking cross-attention in disentangling multi-scale temporal dynamics for robust recognition. A more comprehensive spectral analysis based on joint-wise frequency decomposition is provided in Appendix D.3.

## 5 Conclusion

We propose a novel Signal-SGN, a SNN-based framework designed to learn dynamic features in both the temporal and frequency domains for skeleton-based action recognition. The model comprises multiple stacked 1D-SGC-FSC layers, which utilize SGN to extract temporal-domain features from skeletons converted into spike forms and employ FFT and spiking complex convolution to extract frequency-domain feature. Additionally, the MWTF module, decomposes the spiking-form feature into distinct frequency components for robust feature fusion, enhancing classification performance, significantly. The final model surpasses SOTA SNNs in various aspects and achieves accuracy comparable to traditional ANN models while maintaining extremely low energy consumption, demonstrating its effectiveness and efficiency in skeleton-based action recognition.

## 6 Acknowledgements

The National Natural Science Foundation of China NSFC-61976022 and the BUPT Innovation and Entrepreneurship Support Program 2025-YC-A003 support this work.